\documentclass[10pt,twocolumn,letterpaper]{article}

\usepackage{wacv}              

\usepackage{graphicx}
\usepackage{booktabs}

\usepackage{graphicx}
\usepackage{amsmath}
\usepackage{amssymb}
\usepackage{booktabs}
\usepackage{bm}
\usepackage{stmaryrd}
\usepackage{multirow}
\usepackage{multicol}
\usepackage{mathtools}
\usepackage[ruled,vlined]{algorithm2e}
\usepackage{setspace}
\usepackage{subcaption}

\usepackage{multibib}
\newcites{main}{References}
\newcites{appx}{References}

\DeclareMathOperator*{\argmin}{argmin}
\DeclareMathOperator*{\argmax}{argmax}

\usepackage{tabularx}
\usepackage{subcaption}

\SetKwComment{PythonStyleComment}{\# }{}

\SetCommentSty{mycommfont}
\usepackage {tablefootnote}

\SetKwFor{SimpleRepeat}{repeat}{do}{endrepeat}

\SetKw{Goto}{goto}

\SetKwInput{KwHyperParam}{Hyper params}

\newcommand{\bez}{b\'{e}zier }

\usepackage[pagebackref,breaklinks,colorlinks]{hyperref}

\usepackage[capitalize]{cleveref}
\crefname{section}{Sec.}{Secs.}
\Crefname{section}{Section}{Sections}
\Crefname{table}{Table}{Tables}
\crefname{table}{Tab.}{Tabs.}

\def\wacvPaperID{2305} 
\def\confName{WACV}

\begin{document}

\title{Adversarial Doodles: Interpretable and Human-Drawable Attacks \\ Provide Describable Insights}

\author{Ryoya Nara, Yusuke Matsui\\
The University of Tokyo, Japan\\
{\tt\small \{nara,matsui\}@hal.t.u-tokyo.ac.jp}
}
\maketitle

\begin{abstract}
DNN-based image classifiers are susceptible to adversarial attacks. Most previous adversarial attacks do not have clear patterns, making it difficult to interpret attacks' results and gain insights into classifiers' mechanisms. Therefore, we propose Adversarial Doodles, which have interpretable shapes. We optimize black \bez curves to fool the classifier by overlaying them onto the input image. By introducing random affine transformation and regularizing the doodled area, we obtain small-sized attacks that cause misclassification even when humans replicate them by hand. Adversarial doodles provide describable insights into the relationship between the human-drawn doodle's shape and the classifier's output, such as ``When we add three small circles on a helicopter image, the ResNet-50 classifier mistakenly classifies it as an airplane.'' 
\end{abstract}

\section{Introduction}
\label{sec:intro}

Recent research has revealed that image classification models based on deep neural networks (DNNs) are vulnerable to adversarial attacks~\cite{AE}. Adversarial attacks are deliberate modifications of input images to cause misclassifications. Exploring adversarial attacks is essential for practical applications and theoretical understanding of the classifiers.

Previous studies on adversarial attacks~\cite{fgsm,AE,autoattack} have primarily focused on the effectiveness and imperceptibility of the attacks, rather than their interpretability by humans. As a result, even when an attack successfully fools a classifier, it is difficult to gain insights into the reasons for the model's misclassification. Some studies~\cite{robust-interpretable-attack,why-the-failure,towards-interpretable-attack} have demonstrated a link between interpretability and adversarial attacks. Still, few works have focused on generating interpretable attacks that shed light on the relationship between their attacks and the classifier's outputs.

\begin{figure}[tb]
    \centering
    \begin{minipage}{0.99\hsize}
        \centering
        \includegraphics[width=\linewidth]{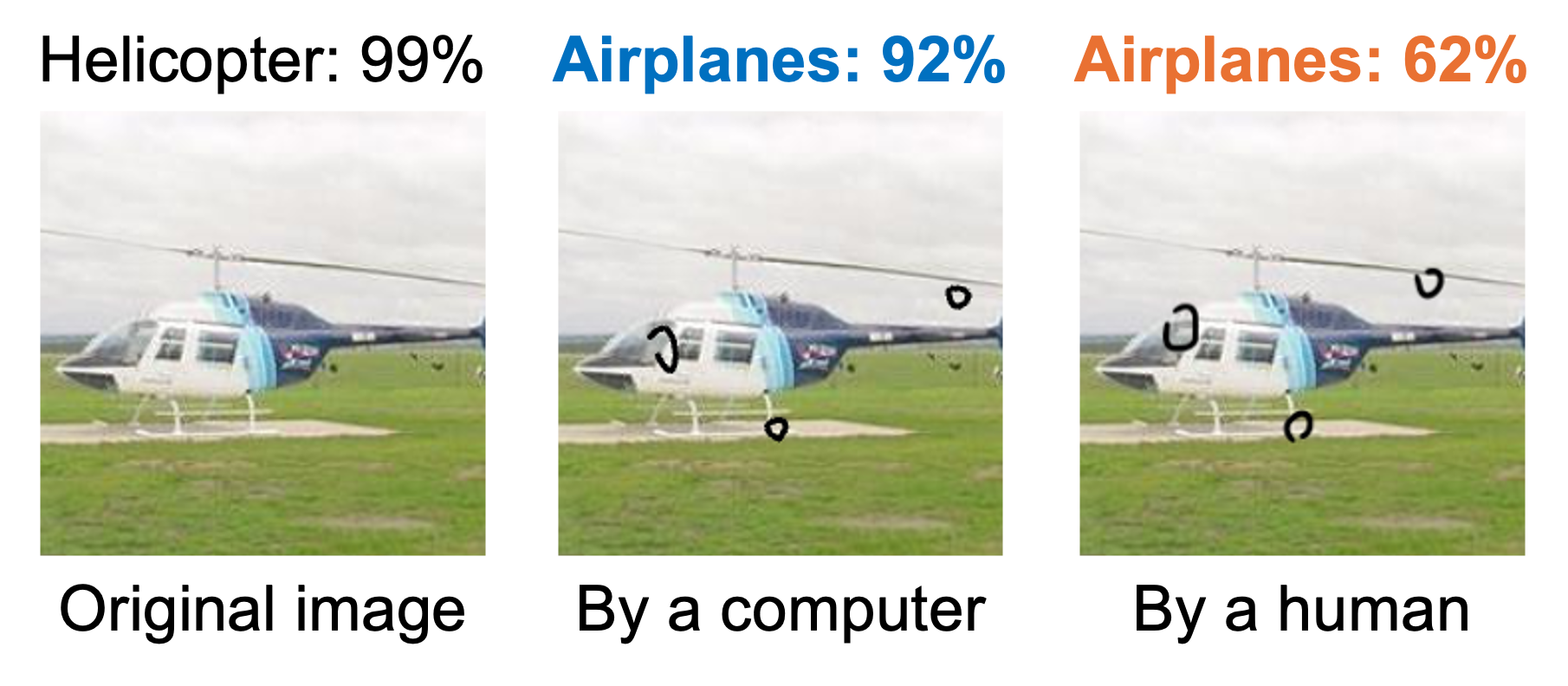}
        \subcaption{Replication examples.}
        \label{fig:title-replication}
    \end{minipage}
    \begin{minipage}{0.99\hsize}
        \centering
        \includegraphics[width=\linewidth]{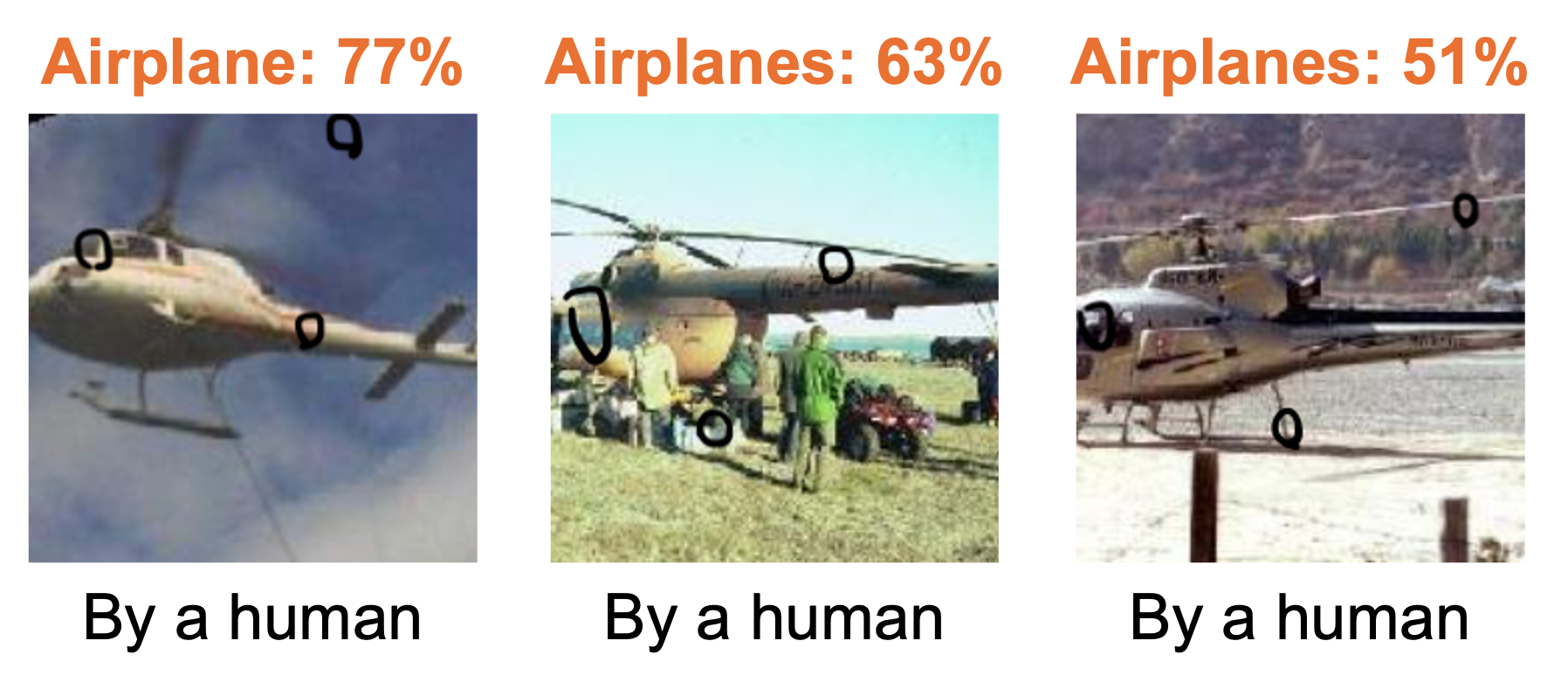}
        \subcaption{Attacks on other images.}
        \label{fig:title-other-images}
    \end{minipage}
    \caption{Examples of our adversarial doodles. Throughout this paper, ``By a computer'' means adding attacks generated by a computer onto the image, and ``By a human'' means adding attacks drawn by a human onto the image. We display an output class of the target classifier with a confidence score above each image. For the character color of each confidence score, we use orange when humans draw attacks on the image, blue when we add a computer-generated attack, and black when we add no attack.}
    \label{fig:title-figure}
\end{figure}

To obtain interpretable adversarial attacks, we propose human-drawable attacks named \textit{adversarial doodles}. We achieve this by optimizing a set of \bez curves to fool the classifier when we overlay them onto an input image. After the optimization, humans can replicate these optimized \bez curves by hand onto the input image and fool the classifier as well.
As simply optimizing \bez curves is insufficient for creating effective doodles, we propose two techniques. First, we introduce random affine image transformations during the optimization. When a human replicates computer-generated adversarial attacks by hand, there must be some misalignment between the computer-generated and human-replicated attacks. To make our attack robust to this misalignment, we simulate the misalignment by applying a random affine transformation to \bez curves during the optimization. Second, we regularize the doodled area to make our attacks' size smaller. When our attack is too large and drastically changes the image's semantics, it becomes self-evident to the human eye that our attack fools the classifier. To avoid this issue, we minimize the doodled area by adding an L1 regularization term for the attack's size to the loss function.

\cref{fig:title-figure} demonstrates adversarial doodles that fool the ResNet-50 classifier pretrained on Caltech-101 dataset. The computer-generated adversarial doodle fools the classifier, as shown in the middle of \cref{fig:title-replication}. Next, a human subject replicates the computer-generated attack, and this human-replicated attack also fools the classifier, as shown in the right of \cref{fig:title-replication}. Furthermore, adversarial doodles discover describable insights into the classifier's mechanism. In \cref{fig:title-figure}, for example, we can interpret the shapes of \cref{fig:title-replication} and get insight: ``When we add three small circles on a helicopter image, the ResNet-50 classifier mistakenly classifies it as an airplane.'' We draw strokes on other helicopter images following this insight, and as a result, we successfully fool the classifier, as shown in \cref{fig:title-other-images}. In other words, adversarial doodles have the potential to provide such describable insights into the relationship between a human-drawn doodle's shape and the classifier's output.

Our goal is to generate interpretable and human-drawable attacks against the classifier. To our knowledge, we are the first to attack image classifiers by human-drawn strokes. As human-drawn strokes can take flexible shapes, adversarial doodles reveal descriable and intriguing insights into the relationship between an attack's shape and the classifier's output. Additionally, since we can easily draw doodles to the objects with pens or sprays, this attack could become a means of physical adversarial attacks~\cite{roadsigns17,advlb}, which directly add attacks to objects in the physical world. Investigating the potential of human-drawn doodles as adversarial attacks is valuable in understanding DNN-based classifiers' mechanisms and exploring practical adversarial attacks.

The main contributions of this paper are as follows:
\begin{itemize}
    \item 
    We propose a method to generate attacks with a set of \bez curves through a gradient-based optimization.
    \item 
    We show that random affine transformation during the optimization enhances the attack success rates when humans replicate the computer-generated attacks. 
    \item
    We observe and interpret adversarial doodles and find some describable insights into the relationship between a human-drawn doodle's shape and the classifier's output.
\end{itemize}

\section{Related Work}

    \subsection{Interpretability of Adversarial Examples}
    Several studies have explored the intersection of adversarial attacks and interpretability. Casper \etal~\cite{robust-interpretable-attack} and Mu \etal~\cite{compositional-explanations} propose ``copy-paste'' attacks where one natural image pasted inside another leads to the model's misclassification, and discuss the results of these attacks to interpret DNN's mechanism. Wang \etal~\cite{generating-semantic-adv} and Zhao \etal~\cite{generating-natural-adversarial-examples} add attacks to latent space and generate semantic adversarial examples to cause misclassification. Tsipras \etal~\cite{robust-interpretable-attack} proposes that the representation of a classifier tends to align better with human perception when we train the classifier to be robust against adversarial examples. Of all of them, Casper \etal~\cite{robust-interpretable-attack} set the most similar goal to ours, which directly leverages perceptible adversarial attack as a tool to obtain insights into DNN classifiers' mechanism.

    \subsection{Shapes of Adversarial Attacks}
    Several studies propose adversarial attacks with simple shapes, for example, one-pixel~\cite{one-pixel-attack}, a patch~\cite{Adversarial-Patch,LaVan}, a triangle~\cite{shadows-can-be-dangerous}, a straight line~\cite{advlb}. Giulivi \etal~\cite{adversarial-scratch} proposes \textit{Adversarial Scratch}, which fools the classifier with a set of \bez curves. 
    The main difference between their work~\cite{adversarial-scratch} and our work is that we make our attack effective even when drawn by humans. We utilize our attacks as interpretable tools to gain insights into the classifiers' mechanism. 

    \subsection{Adversarial Attack Replicated by Humans}
    Several studies focus on adversarial attacks that induce misclassification even when replicated by humans, especially for physical adversarial attacks~\cite{roadsigns17,advlb,transferable-physical-3d-adv}. 
    Physical adversarial attacks consider the scenario where attackers do not have access to image data, and aim to fool the model by directly adding attack onto objects.
    They cause misclassification by printing or pasting adversarial examples~\cite{adversarial-tshirt,Adversarial-Patch}, using a laser beam~\cite{advlb} or a shadow~\cite{shadows-can-be-dangerous}. They enhance the attacks' robustness by introducing random transformation~\cite{syn-robust-adv-ex}, hindering non-printable color~\cite{nps}, or training digital-to-physical transformations~\cite{d2p}. 
    
    We generate adversarial attacks that fool the classifier even when replicated by humans. We add our attack in the digital domain, but get inspiration from EOT~\cite{syn-robust-adv-ex}, which is commonly used in physical adversarial attacks.

\section{Preliminaries}
In this section, we introduce existing technologies that are necessary to implement our proposed approach described in \cref{sec:approach}. 
    \subsection{B\'{e}zier Curves}\label{subsec:bezier-curve}
    A \bez curve is a polynomial curve defined by multiple control points. 
    Coordinates of $N$ control points determine the form of a \bez curve. Given a parameter $x \in [0, 1]$ and coordinates of control points $\{\bm{P}_n\}_{n=1}^{N}$, where each $\bm{P}_n \in \mathbb{R}^2$, we can define the equation of a \bez curve $\bm{B}: [0, 1] \to \mathbb{R}^2$ as \cref{eq:bezier}~\cite{BezierCurveEOM}. 
    \begin{equation}\label{eq:bezier}
        \bm{B}(x; \bm{P}_1,\ldots \bm{P}_N) = \sum_{n=1}^{N} \binom{N-1}{n-1} x^{n-1}(1-x)^{N-n}\bm{P}_n.
    \end{equation}
    Here, $\binom{\cdot}{\cdot}$ denotes the binomial coefficient.

   Generating a \bez curve by optimizing the coordinates of control points is more straightforward than generating a raster image of the curve. Therefore, previous studies about sketch generation~\cite{sketchGen,bezier-sketch,cloud2curve} approximate sketch strokes with \bez curves and optimize their control points to obtain the desired sketches. We also follow this line and represent an adversarial doodle by a set of \bez curves.

    \subsection{Differentiable Rasterizer}
    
    Li \etal propose \textit{differentiable rasterizer}~\cite{differentiable-vector-graphic}, a function that maps the parameters of a vector image to a raster image. With this innovation, we can apply machine learning techniques, such as backpropagation-based optimization with stochastic gradient descent, to vector images.
    
    Since then, several studies about sketch generation~\cite{clipdraw,clipascene,clipasso} and SVG generation~\cite{svgdreamer} have leveraged the differential rasterizer module to optimize vector graphics in a gradient-based way. To generate human-drawable attacks, we follow these works and use the differentiable rasterizer to optimize \bez curves.

\section{Approach}\label{sec:approach}


We generate human-drawable and interpretable attacks by optimizing the coordinates of \bez curves' control points. Here, we require robustness against the misalignment between strokes generated by a computer and those replicated by a human. Furthermore, our attacks should not change the appearance of input images drastically. Therefore, our attack's size, or the ratio of the doodled area in an image, should be as small as possible. 

We consider the white-box and untargeted attack problem. We aim to get insights into the relationship between the shapes of doodles and the classifier's output, and run the white-box attack to obtain powerful attacks. We focus on the untargeted attack, where our goal is simply to induce misclassification and do not care about the resulting class.

\subsection{Formulation}\label{subsec:problem-settings}
We seek adversarial doodles that, when overlaid on the input image, cause the classifier's misclassification. The input image is denoted as $\bm{X} \in \mathbb{R}^{H \times W \times 3}$, which has the height $H$ and the width $W$ and three channels. Our study focuses on the $c$-class classification problem. We define the classifier as $\bm{f}: \mathbb{R}^{H \times W \times 3} \to \mathbb{R}^c$ and denote the $i$\textsuperscript{th} element of $\bm{f}$ as $f_i$, which represents the confidence score of the $i$\textsuperscript{th} class ($i \in \llbracket c \rrbracket$)\footnote{For $N \in \mathbb{N}$, we define  $\llbracket N \rrbracket$ as $\{1, 2, \dots, N\}. $}. The classifier's output is the class whose confidence score is the highest: $\argmax_{x \in \llbracket c \rrbracket}f_x(\bm{X})$.
The ground-truth class of $\bm{X}$ is $s \in \llbracket c \rrbracket$.
We aim to make the output of $\bm{f}$ apart from $s$.
    
In this paper, we define an adversarial doodle as a set of $L$ \bez curves with $N$ control points each. The \bez curves have a black color and a fixed width. We stack all the coordinates of their control points and parameterize an adversarial doodle as $\bm{V} \in \mathbb{R}^{L \times N \times 2}$. For $l \in \llbracket L \rrbracket$ and $n \in \llbracket N \rrbracket$, the $(l, n)$ element of $\bm{V}$ is denoted as $\bm{P}_{l, n} \in \mathbb{R}^2$, which means the coordinate of $l$\textsuperscript{th} \bez curve's $n$\textsuperscript{th} control point.  We fix $L$ and $N$, and optimize the parameter $\bm{V}$ to find the most powerful adversarial doodle.

Additionally, we introduce two mathematical functions to accurately formulate the problem. First, we define a distribution $\mathcal{T}$ of a random affine transformation for a binary image $\bm{t}: \{0, 1\}^{H \times W} \to \{0, 1\}^{H \times W}$. We use $\mathcal{T}$ to simulate the misalignment occurring when humans replicate the computer-generated attack. This technique is motivated by EOT~\cite{syn-robust-adv-ex}, which is commonly used to generate robust attacks in physical adversarial attack scenarios~\cite{advlb,syn-robust-adv-ex,shadows-can-be-dangerous}. 
Next, we define a rasterizer $\bm{\phi}: \mathbb{R}^{L \times N \times 2} \to \{0, 1\}^{H \times W}$ which converts the coordinates of \bez control points into raster information of \bez curves. Here, $\bm{\phi}(\bm{V})$ is a rendered doodle and represents if the attack exists in each pixel. For $h \in \llbracket H \rrbracket, w \in \llbracket W \rrbracket$, if the $(h, w)$ element of $\bm{\phi}(\bm{V})$ equals $1$, it means that the $(h, w)$ pixel is attacked. 

Using the functions defined above, we now define the final pixel-wise doodling operation $\bm{\psi}: \mathbb{R}^{H \times W \times 3} \times \mathbb{R}^{L \times N \times 2} \to \mathbb{R}^{H \times W \times 3}$. $\bm{\psi}$ expresses the operation of drawing \bez curves onto the input image $\bm{X}$ with random transformation $\bm{t}$ as \cref{eq:sketch-blend}.
\begin{equation}
    \label{eq:sketch-blend} \bm{\psi}(\bm{X}, \bm{V}; \bm{t}) \coloneq \bm{X} \odot (\mathbf{1} - \bm{t}(\bm{\phi}(\bm{V}))). 
\end{equation}
Here, $\mathbf{1}$ denotes a $H \times W$ matrix whose elements are all 1, and $\odot$ denotes the element-wise product.\footnote{Similar to many papers, we also abuse this operator slightly; $\bm{\bm{X}}\odot\bm{(\mathbf{1} - \bm{t}(\bm{\phi}(\bm{V})))}$ computes the element-wise product between $\bm{(\mathbf{1} - \bm{t}(\bm{\phi}(\bm{V})))}$ and each channel of $\bm{X}$. }
Note that we apply the transformation $\bm{t}$ only to a set of \bez curves, not to the image attacked by \bez curves.

We aim to find the smallest-sized adversarial doodles that cause the misclassification even with any random affine transformation. Here, a doodle's size can be formulated as $\frac{\Vert \bm{\phi}(\bm{V}) \Vert_1}{HW}$, which means the doodled area's ratio after rasterization compared with the input image size. Now, the optimal attack $\bm{V}^* \in \mathbb{R}^{L \times N \times 2}$ can be formulated as \cref{eq:optimization}:
\begin{align}\label{eq:optimization}
    & \bm{V}^* = \argmin_{\bm{V} \in \mathbb{R}^{L \times N \times 2}}\Vert \bm{\phi}(\bm{V}) \Vert_1 \\
    & \text{subject to~} s \ne \argmax_{x \in \llbracket c \rrbracket} f_x(\bm{\psi}(\bm{X}, \bm{V}; \bm{t}))  ~~~ \forall \bm{t} \sim \mathcal{T} \notag
\end{align}
Here, $s \ne \argmax_{x \in \llbracket c \rrbracket} f_x(\bm{\psi}(\bm{X}, \bm{V}; \bm{t}))$ means that the classifier's output must be different from the ground-truth class when we add the attack onto the input image with a random affine transformation $\bm{t}$. The condition $\forall \bm{t} \sim \mathcal{T}$ indicates that $\bm{V}$ must fool the classifier whatever random small affine transformation we apply, meaning the doodle is robust against small misalignment.

\subsection{Optimization}

\paragraph{Overview}

\begin{figure}
    \centering
    \includegraphics[width=0.98\linewidth]{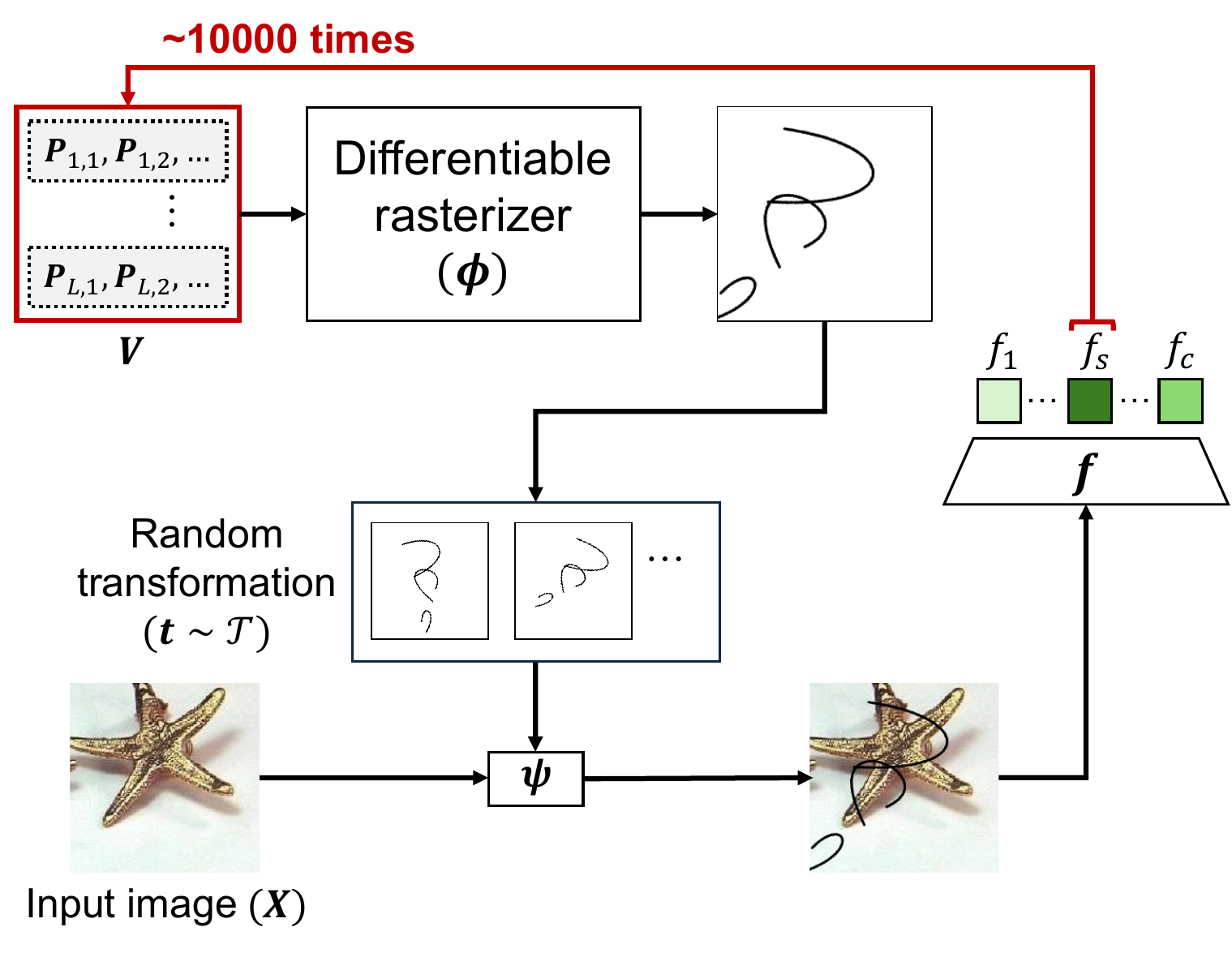}
    \caption{Overview of our proposed method.}
    \label{fig:proposed_method}
\end{figure}

\cref{fig:proposed_method} illustrates the overall framework of our proposed method to obtain the optimal adversarial doodles described in \cref{eq:optimization}. 
We initialize $\bm{V}$ and draw doodles on the image $\bm{X}$ with a random affine transformation $\bm{t}$. The doodled image $\bm{\psi}(\bm{X}, \bm{V}; \bm{t})$ is provided to the classifier $\bm{f}$. We repeatedly update $\bm{V}$ with a gradient-descent way to make the confidence score of the ground-truth class $f_s$ smaller.
Our purpose in introducing random affine transformation $\bm{t} \sim \mathcal{T}$ is to enhance the robustness against the misalignment. Additionally, to enable this gradient-based optimization, we use a differentiable rasterizer~\cite{differentiable-vector-graphic} as $\bm{\phi}$. This allows us to perform the normally non-differentiable operation of rasterization differentially.

\paragraph{Algorithm}

\begin{algorithm}[tb]
    \DontPrintSemicolon
    \caption{Psudocode of our proposed method.}
    \label{alg:optimization}
    \setstretch{1.1}
    \nl\KwIn{$\bm{X} \in \mathbb{R}^{H \times W \times 3}, ~ s \in \llbracket c \rrbracket$} 
    \nl\KwOut{$\bm{V}^* \in \mathbb{R}^{L \times N \times 2}$}
    \nl\KwHyperParam{$B \in \mathbb{N}$, ~ $N_\mathrm{itr} \in \mathbb{N}$, ~ $\alpha \in \mathbb{R}_{> 0}$}
    \nl$\bm{V}, \bm{V}^* \leftarrow \mathrm{Initialize~with~random~values}$\\
    \nl$S_\mathrm{min} \leftarrow \infty$ \PythonStyleComment*[r]{Size of the best doodle}
    \nl \SimpleRepeat{$N_\mathrm{itr}~\mathrm{times}$}{
        \nl \PythonStyleComment*[l]{(1) Training phase}
        \nl$\mathcal{L} \leftarrow 0$ \PythonStyleComment*[r]{Loss}
        \nl\SimpleRepeat{$B~\mathrm{times}$}{
            \nl$\bm{t} \gets \mathrm{Draw~from}~\mathcal{T}~\mathrm{randomly}$ \\
            \nl$\mathcal{L} \leftarrow \mathcal{L} + \log{(f_s(\bm{\psi}(\bm{X}, \bm{V}; \bm{t})))}$
        }
        \nl$\mathcal{L} \leftarrow \frac{\mathcal{L}}{B}$\\
        \nl\If(\PythonStyleComment*[f]{Start to shrink}){$S_\mathrm{min} \ne \infty$}{
            \nl$\mathcal{L} \leftarrow \mathcal{L} + \alpha \frac{\Vert \bm{\phi}(\bm{V})\Vert_1}{HW}$
        }
        \nl$\bm{V} \gets \mathrm{Update}(\mathcal{L}, \bm{V})$ \PythonStyleComment*[r]{Back propagate}
        \;   
        \nl \PythonStyleComment*[l]{(2) Validation phase}
        \nl\SimpleRepeat{$B~\mathrm{times}$}{
            \nl$\bm{t} \gets \mathrm{Draw~from}~\mathcal{T}~\mathrm{randomly}$ \\
            \nl\If{$s = \argmax_{x \in \llbracket c \rrbracket} f_x(\bm{\psi}(\bm{X}, \bm{V}; \bm{t}))$}{
                \nl \PythonStyleComment*[l]{Fails. Train again.}
                \nl \Goto{$\mathrm{L7}$}\\
            }
        }      
        \nl\If{$\frac{\Vert \bm{\phi}(\bm{V})\Vert_1}{HW} \leq S_\mathrm{min}$}{
            \nl \PythonStyleComment*[l]{Maintain the best doodle.}
            \nl$\bm{V}^* \leftarrow \bm{V}$\\
            \nl$S_\mathrm{min} \leftarrow \frac{\Vert \bm{\phi}(\bm{V})\Vert_1}{HW}$ 
        }
    }
    \nl\Return{$\bm{V}^*$}
\end{algorithm}

Although \cref{fig:proposed_method} may appear simple, we incorporate various strategies. \cref{alg:optimization} describes the details of our proposed algorithm. The inputs are an image $\bm{X}$, and its ground-truth class $s$. The output is the optimal attack $\bm{V}^*$. First, we initialize $\bm{V}$ and $\bm{V}^*$ (L4). Next, we prepare $S_{\mathrm{min}}$ to record the temporary minimum doodle size (L5). After that, we update $\bm{V}$ $N_{\mathrm{itr}}$ times (L6). Each iteration consists of a training phase and a validation phase.

In the training phase, we update $\bm{V}$ to fool the classifier. First, we choose a random affine transformation $\bm{t}$ (L10) and obtain a doodled image $\bm{\psi}(\bm{X}, \bm{V}; \bm{t})$. Next, we introduce a positive cross-entropy loss ($\log{(f_s(\bm{\psi}(\bm{X}, \bm{V}; \bm{t})))}$) to fool the classifier. We repeat this $B$ times to make our attack robust to various misalignments (L8-12). After calculating the loss, we add the L1 regularization term to the loss to make the attack's size smaller (L14). The \textbf{if} statement in L13 is crucial for the success of the overall optimization. We first skip this \textbf{if} statement. That is, we optimize $\bm{V}$ without considering the attack's size (L14) until the attack succeeds later in L25. Once the attack succeeds, the \textbf{if} statement in L13 is activated, and we start to minimize the attack's size as much as possible.

In the validation phase, we confirm if the obtained $\bm{V}$ can fool the classifier and update the best attack $\bm{V}^*$ to obtain the smallest attack. First, we check if the obtained $\bm{V}$ successfully causes the misclassification even when we add various random transformations $t$ (L17-21). If $\bm{V}$ passes the check, we calculate the obtained attack's size $\frac{\Vert \bm{\phi}(\bm{V})\Vert_1}{HW}$. If the size is smaller than the current smallest attack's size $S_{\mathrm{min}}$, we update $\bm{V}^*$.

Overall, our algorithm is based on two important designs: (1) Our algorithm trains $\bm{V}$ until the attack successfully fools the classifier. (2) Once the attack succeeds, we reduce the attack's size.
If we try to fool the classifier and reduce the attack's size simultaneously, we often fail to fool the classifier even after optimizing $\bm{V}$ many times.

\section{Experiments}\label{sec:experiments}

\begin{figure}[tb]
    \centering
    \includegraphics[width=0.98\linewidth]{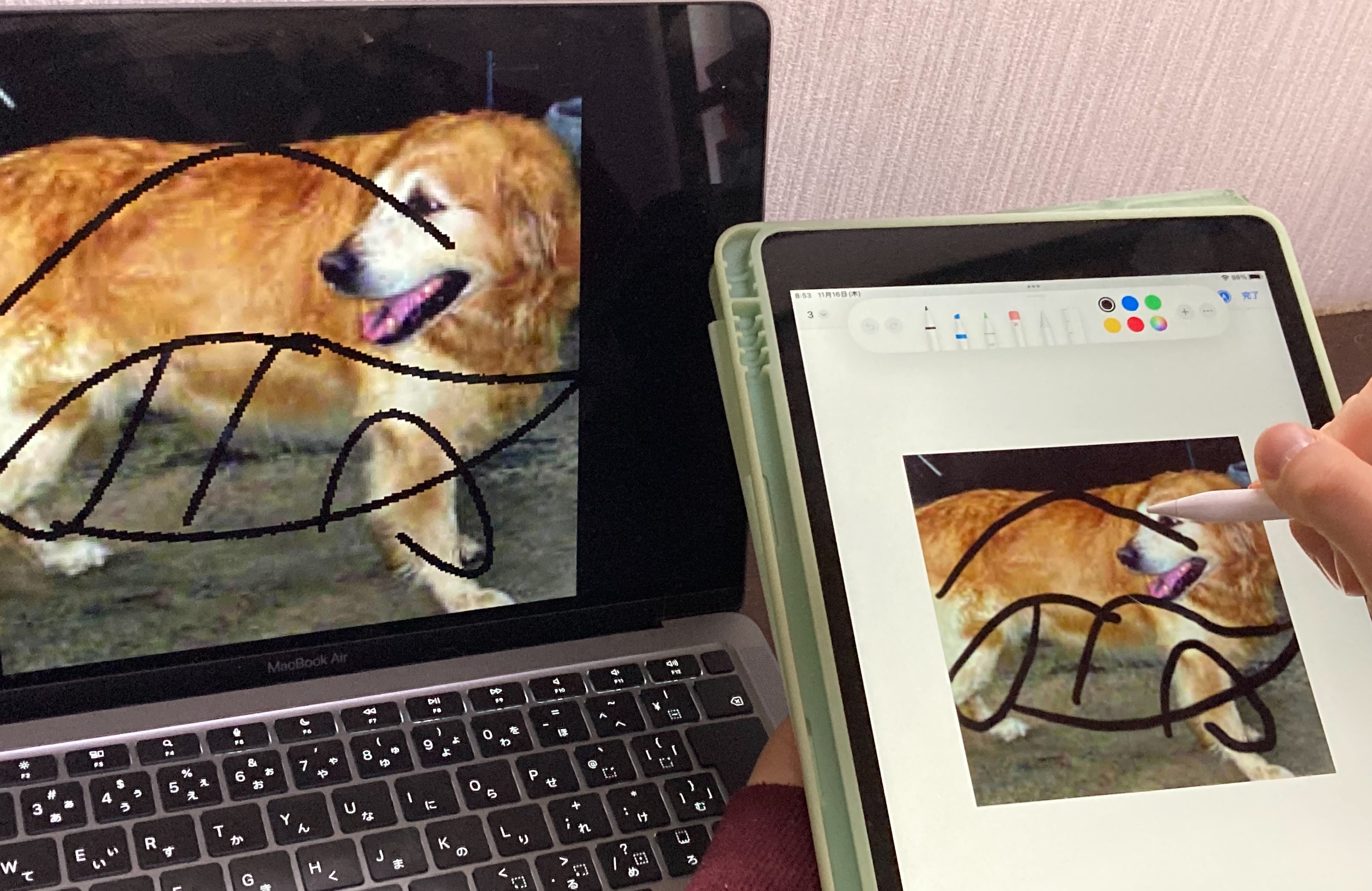}
    \caption{Human replication settings with a tablet. A human subject displays an adversarial doodle optimized by a computer on the PC's screen and replicates it to draw black strokes with a tablet.}
    \label{fig:ipad}
\end{figure}

We run experiments to confirm if we can generate human-drawable adversarial attacks through our proposed method. Our experiments have two steps. In the first step, we evaluate computer-generated adversarial doodles: For each image, we generate an attack $\bm{V}^*$ via \cref{alg:optimization} and confirm if the attack generated on a computer successfully fools the classifier. In the second step, we evaluate human-replicated adversarial doodles: we ask the human subjects to replicate attacks generated in the first step with digital devices as shown in \cref{fig:ipad}, and confirm if the human-drawn attacks successfully fool the classifier.

\subsection{Dataset and Classifiers}
Throughout our experiments, we use the Caltech-101 dataset~\cite{caltech-101} and set up a $101$-class classification problem on this dataset. We choose this dataset for two reasons. First, the size of each image is about $200 \times 300$ pixels, which is enough for human subjects to replicate attacks in freehand. Since human subjects draw an attack to each image in our experiments, low-resolution datasets, like MNIST and Cifar-10, are not suitable for our experiments. Second, compared with ImageNet-1K~\cite{imagenet}, the number of the classes is not so large, and the classes are not so similar. Various adversarial attack studies use ImageNet-1K. However, since our studies handle untargeted attack settings, a dataset with too many classes, like ImageNet-1K, could easily result in a successful attack.

We train the ResNet-50 and ViT-B/32 classifiers on Caltech-101 and freeze them as the classifiers to fool in our experiments. Before inputting each image into the classifier, we resize each image to $256 \times 256$ and crop the center of it to $224 \times 224$. We train the classifiers for Caltech-101 based on a standard train-validation pipeline. As a result of the training, the ResNet-50 classifier achieves $92\%$ accuracy, and the ViT-B/32 classifier achieves $95\%$ accuracy. Throughout our experiments, we aim to generate attacks that fool these frozen classifiers.

We set the number of control points for each \bez curve ($N$) to $4$, and the \bez curve's width to $1.5$ px. We experiment with two patterns of cases: one \bez curve ($L=1$) and three \bez curves ($L=3$). We use Adam~\cite{adam} as the optimizer and set the learning rate to $1$, the same value as CLIPAsso~\cite{clipasso}. For hyperparameters described in \cref{alg:optimization}, we set $N_{\mathrm{itr}}=10000, B=10, \alpha=1$. We use a single Tesla V100 GPU for training.

\subsection{Experimental Design}\label{subsec:experimental-design}

In the first step, we evaluate if attacks generated by \cref{alg:optimization} can successfully fool the classifiers. To evaluate this, we randomly collect $400$ images that are successfully classified as the correct class by each classifier. As we collect images separately for ResNet-50 and ViT-B/32, the images we prepare for each classifier are different.

In the second step, we evaluate if attacks generated in the first step can also fool the classifier even when human subjects replicate them in freehand. For each classifier, we collect images where the attack generated in the first step can successfully fool the classifier, and use them in this step. In other words, we exclude images where the attack generated in the first step fails to fool the classifier. Next, we shuffle the collected images and divide them into $20$ subsets. Then, we recruit $20$ human subjects and randomly give one subset to each. We ask the human subjects to replicate the attacks generated in the first step onto the images included in the subsets assigned to themselves. When the human subjects replicate attacks by hand, they use an image editor on PCs or tablets (see \cref{fig:ipad}).

\subsection{Results}

\begin{table}[tb]
    \caption{The number of attack success cases of computer-generated attacks and human-replicated attacks.}
    \label{tab:attack-result}
    \centering
    \begin{tabular}{@{}lllll@{}} \toprule
          & & & \multicolumn{2}{c}{\# attack success cases} \\ \cmidrule(){4-5}
        Architecture & $L$ & \# total images &  Computer & Human \\ \midrule
        ResNet-50& $1$ & $400$ & $64$ & $52$ \\ \cmidrule(){3-5}
        & $3$ & $400$ & $\bm{247}$ & $\bm{208}$ \\ \midrule
        ViT-B/32& $1$ & $400$ & $53$ & $35$ \\ \cmidrule(){3-5}
        & $3$ & $400$ & $\bm{171}$ & $\bm{131}$ \\ \bottomrule
    \end{tabular}
\end{table}

\begin{figure}[tb]
    \centering
    \begin{minipage}{0.98\hsize}
        \begin{minipage}{\hsize}
            \centering
            \includegraphics[width=\hsize]{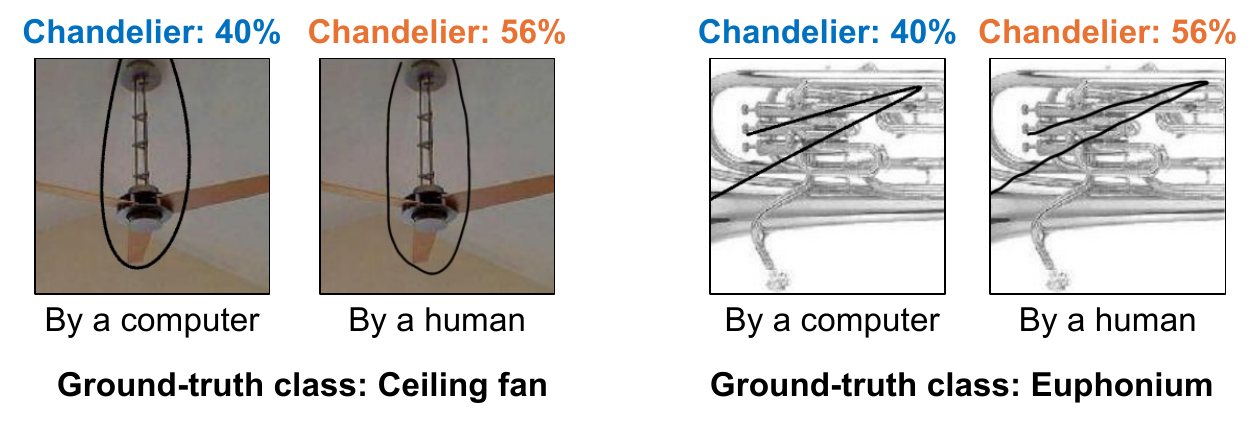}
        \end{minipage}
        \begin{minipage}{\hsize}
            \centering
            \includegraphics[width=\hsize]{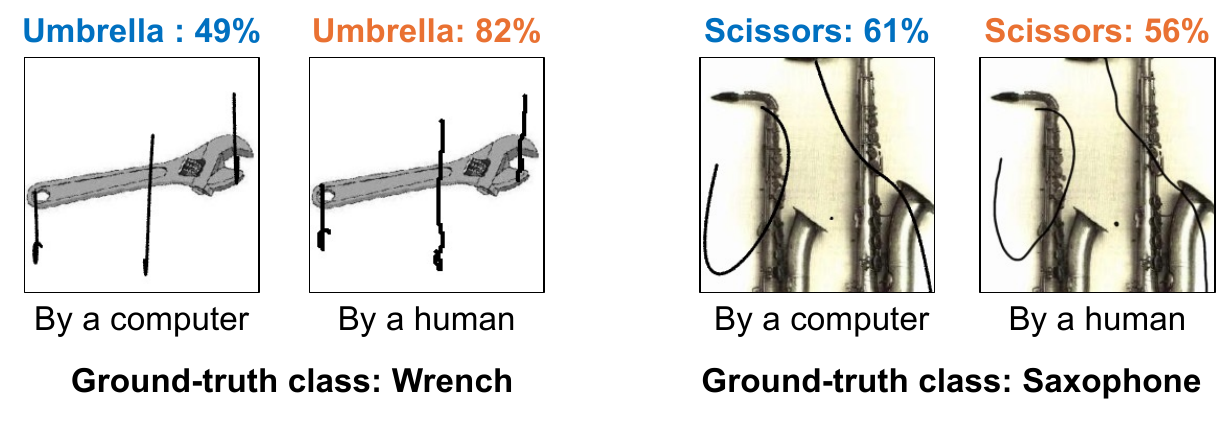}
        \end{minipage}
        \subcaption{Success cases}
        \label{fig:replicated-attack-examples:success}
    \end{minipage}
    \begin{minipage}{0.98\hsize}
        \begin{minipage}{\hsize}
            \centering
            \includegraphics[width=\hsize]{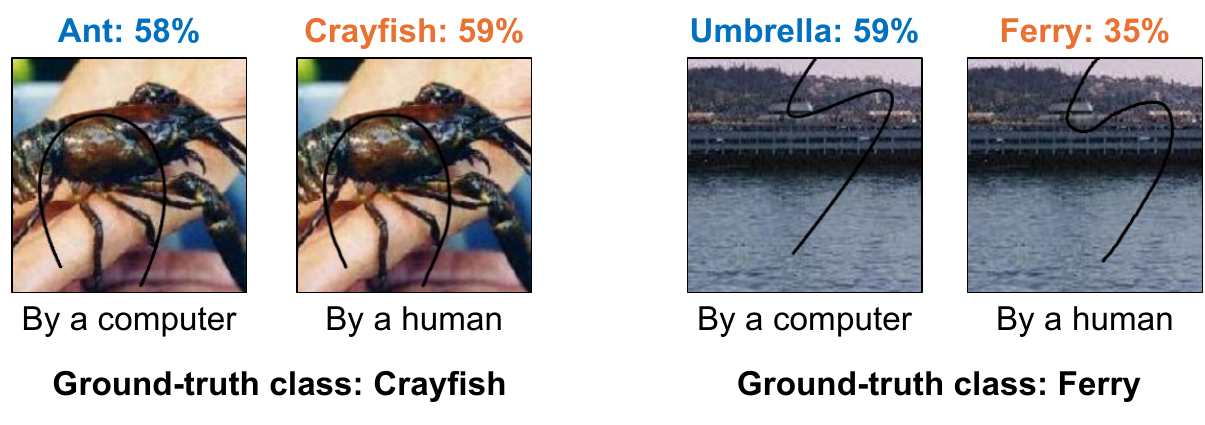}
        \end{minipage}
        \begin{minipage}{\hsize}
            \centering
            \includegraphics[width=\hsize]{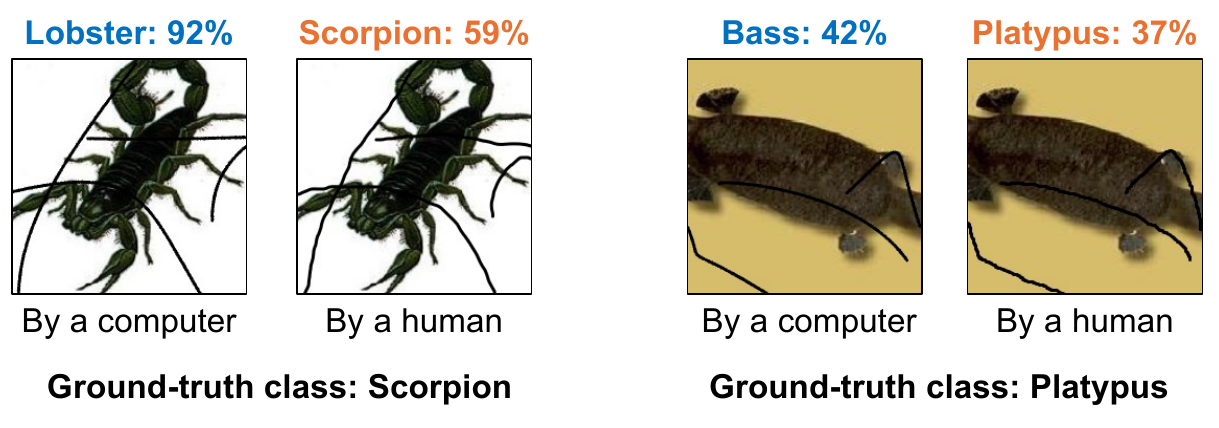}
        \end{minipage}
        \subcaption{Failure cases}
        \label{fig:replicated-attack-examples:failure}
    \end{minipage}
    \caption{Examples of adversarial doodles for the ResNet-50 classifier. \cref{fig:replicated-attack-examples:success} shows the cases where both computer-generated and human-replicated attack successfully fool the ResNet-50 classifier. \cref{fig:replicated-attack-examples:failure} shows the cases where computer-generated attacks successfully fool the classifier, but human-replicated attacks fail.}
    \label{fig:replicated-attack-examples}
\end{figure}

\cref{tab:attack-result} describes the overall results. Attacks with three \bez curves surpass attacks with one \bez curve (e.g., $53$ vs $171$ for ResNet-50, human-replicated), which infers that more \bez curves lead to more attack success rates. In terms of the classifier's architecture, the ViT-B/32 classifier is a little more robust to both computer-generated and human-replicated adversarial doodles than the ResNet-50 classifier (e.g., $35$ vs $52$ for $L=1$, human-replicated).
\cref{fig:replicated-attack-examples} shows examples of adversarial doodles for the ResNet-50 classifier. In some cases, the human-replicated attack successfully fools the classifier; in other cases, the human-replicated attack fails. We analyze these cases more in \cref{subsec:gradCam}.

\begin{figure*}[tb]
    \centering
    \begin{minipage}{0.98\hsize}
        \centering
        \fbox{
            \includegraphics[width=\linewidth]{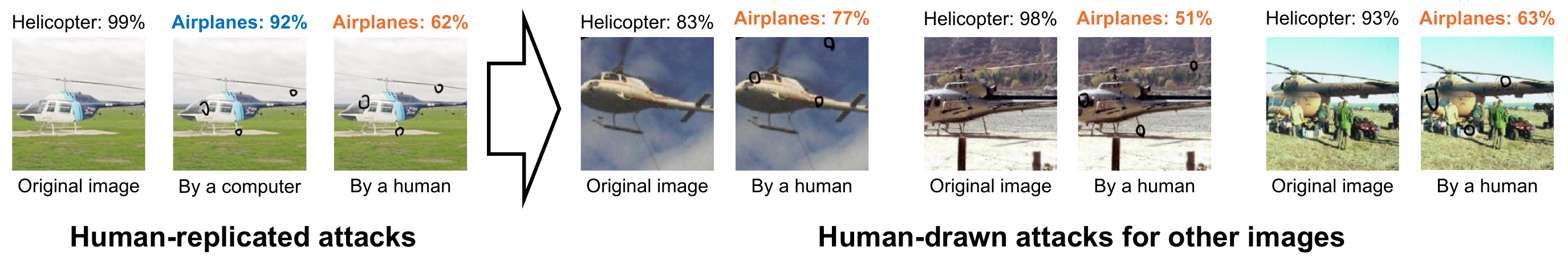}
        }
        \subcaption{Helicopter $\to$ Airplanes: When we draw three small circles on a helicopter image, the image is classified as an airplane. }
        \label{fig:describable-insights:helicopter2airplane}
        \vspace{0.2cm}
    \end{minipage}
    \begin{minipage}{0.98\hsize}
        \centering
        \fbox{
            \includegraphics[width=\linewidth]{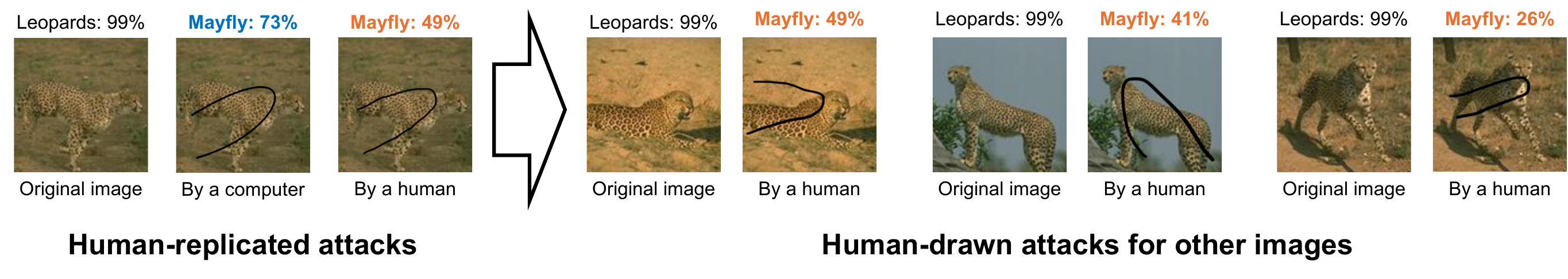}
        }
        \subcaption{Leopards $\to$ Mayfly: When we draw one curve on the leopard’s body in an image, the image is classified as a mayfly.}
        \label{fig:describable-insights:leopards2mayfly}
        \vspace{0.2cm}
    \end{minipage}
    \begin{minipage}{0.98\hsize}
        \centering
        \fbox{
            \includegraphics[width=\linewidth]{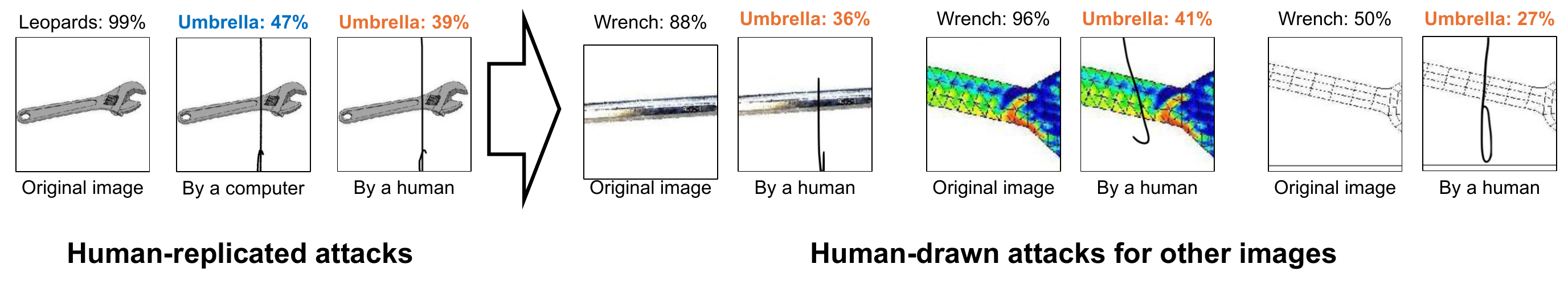}
        }
        \subcaption{Wrench $\to$ Umbrella: When we draw one line on a wrench image, the image is classified as an umbrella. }
        \label{fig:describable-insights:wrench2umbrella}
    \end{minipage}
    \caption{Examples of describable insights into the ResNet-50 classifier's mechanism. The human-replicated attacks shown in this figure are selected from the experiments described in \cref{sec:experiments}. The human-drawn attacks for other images are not the replication of a computer-generated attack. We gain insight from the results of human-replicated attacks and draw doodles on other images following the insight.}
    \label{fig:describable-insights}
\end{figure*}
\section{Describable Insights}

By utilizing adversarial doodles, we find describable insights into the relationship between a human-drawn doodle's shape and the ResNet-50 classifier's output. We observe and interpret the adversarial doodles obtained in our experiments (\cref{sec:experiments}), and gain insights about how to fool the classifier by drawing a simple doodle. After that, we draw an adversarial doodle on other images following the insights and evaluate if the doodle fools the ResNet-50 classifier.

\cref{fig:describable-insights} shows the examples of the insights. For example, in the case of \cref{fig:describable-insights:leopards2mayfly}, we observe and interpret the human-replicated attacks, and extract a describable insight: ``When we draw one curve on the leopard’s body in an image, the image is classified as an airplane.'' We draw doodles on other leopard images following this insight, and as a result, the ResNet-50 classifier classifies each image as ``mayfly.''



\section{Analysis}

\subsection{Random Affine Transformation}\label{subsec:random-affine}
\begin{figure}[tb]
    \centering
    \begin{minipage}{0.49\hsize}
        \centering
        \includegraphics[width=\hsize]{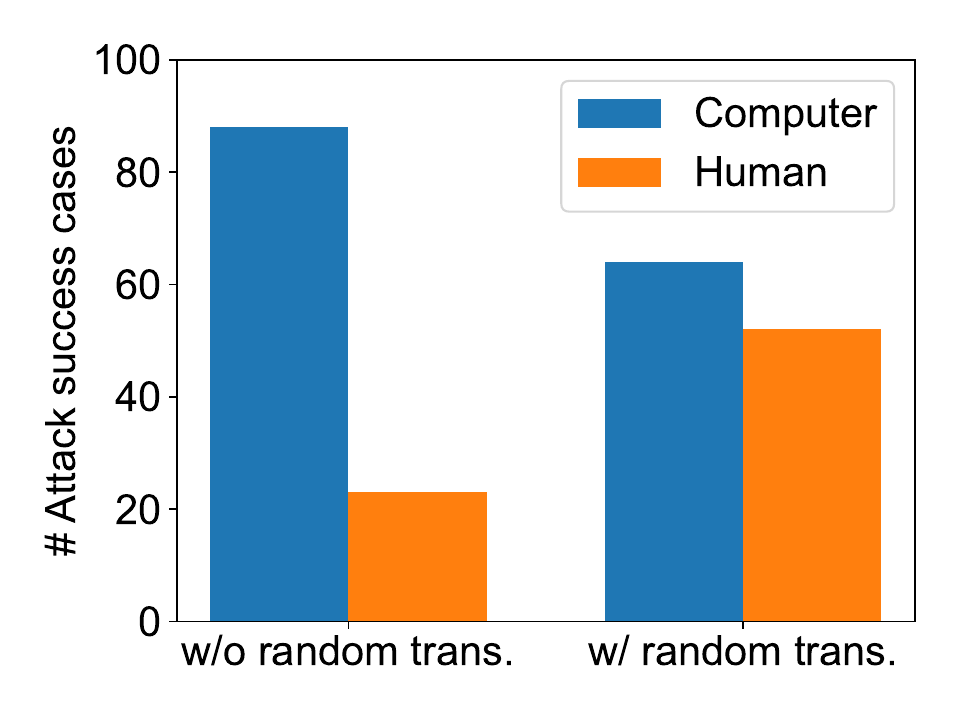}
        \subcaption{$L=1$, ResNet-50}
    \end{minipage}
    \hfill
    \begin{minipage}{0.49\hsize}
        \centering
        \includegraphics[width=\hsize]{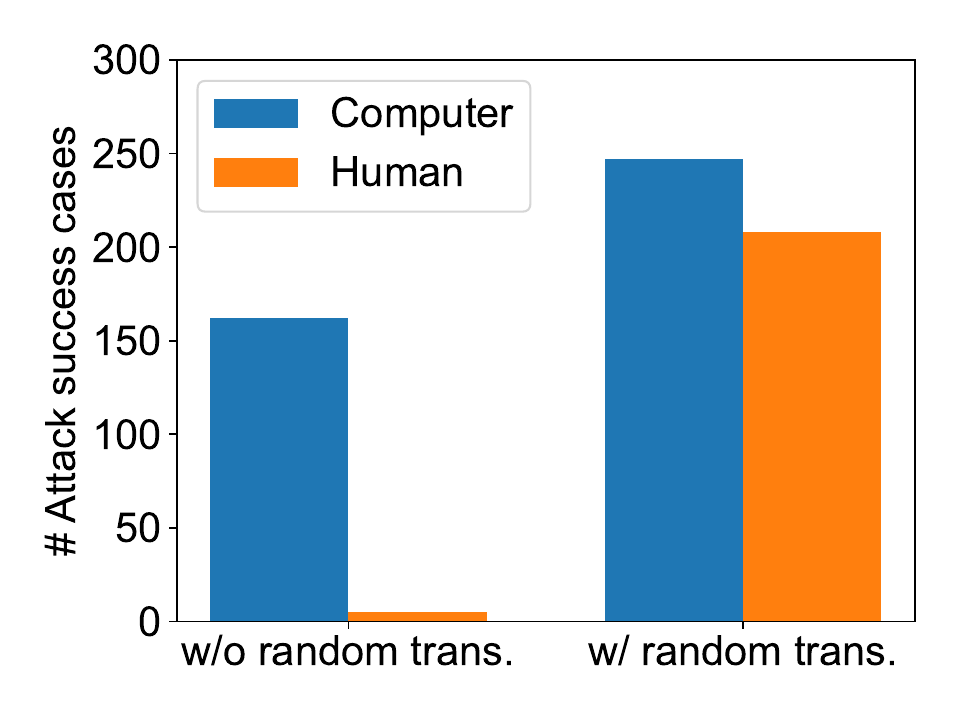}
        \subcaption{$L=3$, ResNet-50}
    \end{minipage}
    \begin{minipage}{0.49\hsize}
        \centering
        \includegraphics[width=\hsize]{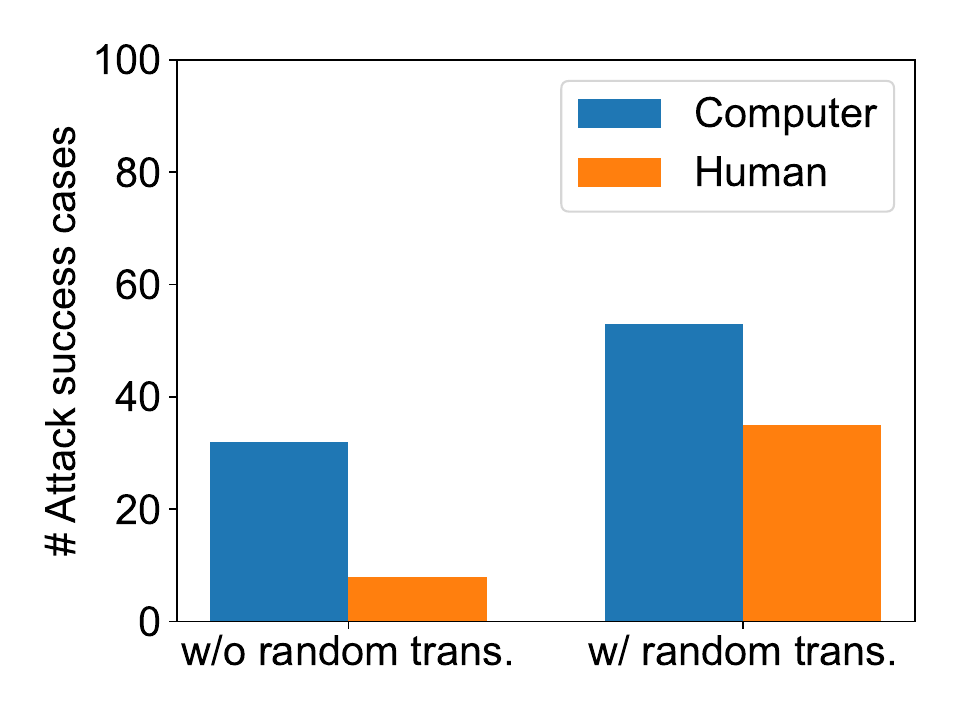}
        \subcaption{$L=1$, ViT-B/32}
    \end{minipage}
    \hfill
    \begin{minipage}{0.49\hsize}
        \centering
        \includegraphics[width=\hsize]{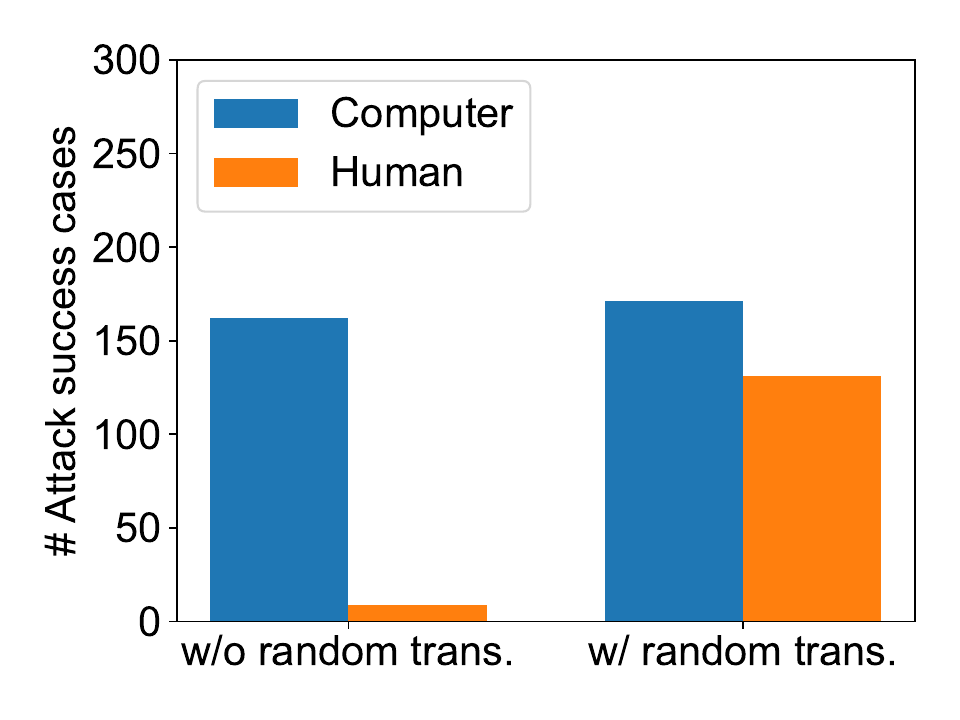}
        \subcaption{$L=3$, ViT-B/32}
    \end{minipage}
    \caption{The relationship between random transformation and the number of attack success cases.}
    \label{fig:ablation-study-results}
\end{figure}

We evaluate whether random affine transformation enhances robustness against the misalignment between generated and replicated doodles. We execute the ablation study in the same settings as \cref{sec:experiments}. In the first step, we generate adversarial doodles described in \cref{sec:experiments} in two settings: (1) adding random transformation during the optimization and (2) no transformation. In the second step, we follow human replication experiments described in \cref{sec:experiments} and evaluate if replicated attacks cause misclassification.

\cref{fig:ablation-study-results} demonstrates that random affine transformation improves the replicated attack success rates while not interfering with optimization on the computer. We can significantly increase the number of human-drawn attack success cases (represented as orange bars in \cref{fig:ablation-study-results}) by introducing random affine transformation. Therefore, random affine transformation works well when considering that humans replicate our attack by hand and fool the classifier.

\subsection{Activation Visualization with GradCAM}\label{subsec:gradCam}
To explain what part of each adversarial doodle leads to misclassification, we analyze the reason for the classifier's decision. We use GradCam~\cite{gradcam} and visualize what area of the image the classifier focuses on. Next, we observe the results and discuss why human-replicated attacks sometimes succeed and sometimes fail.

\begin{figure}[tb]
    \centering
    \begin{minipage}[b]{0.48\hsize}
        \centering
        \includegraphics[width=\hsize]{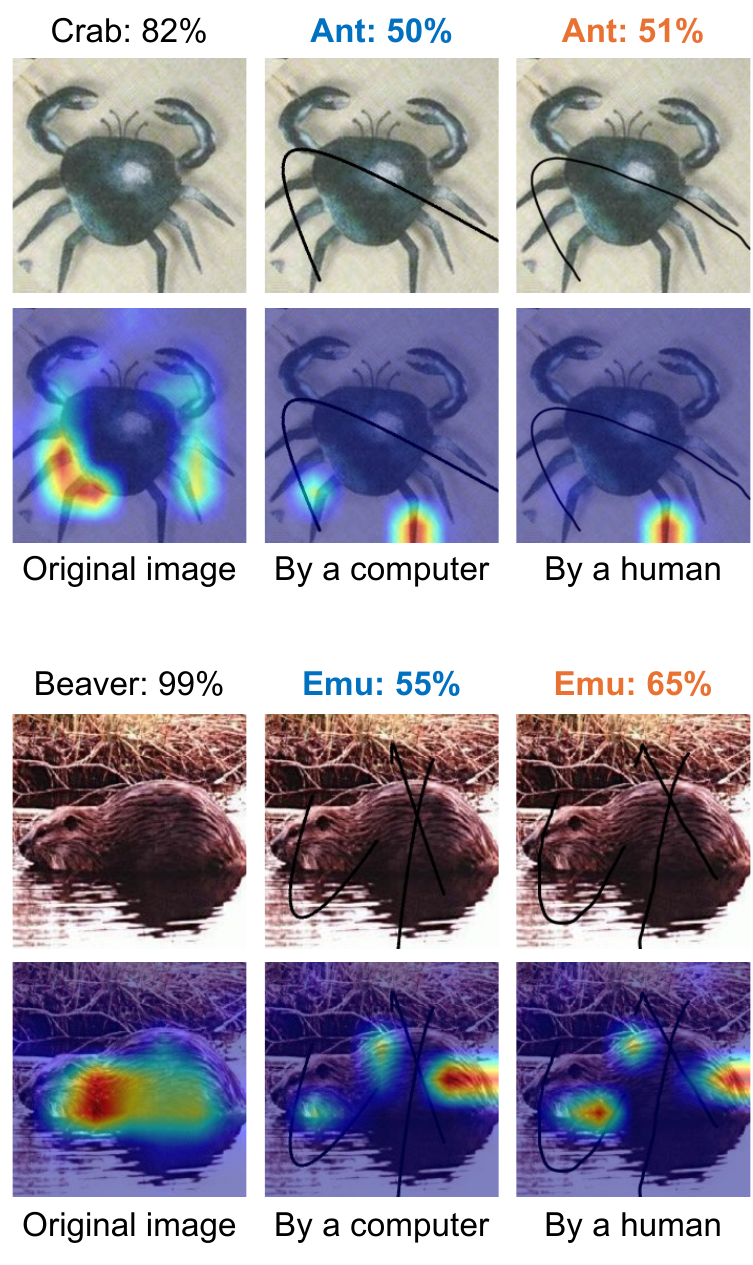}
        \subcaption{Success cases}
        \label{fig:gradcam:success}
    \end{minipage}
    \hfill
    \begin{minipage}[b]{0.49\hsize}
        \centering
        \includegraphics[width=\hsize]{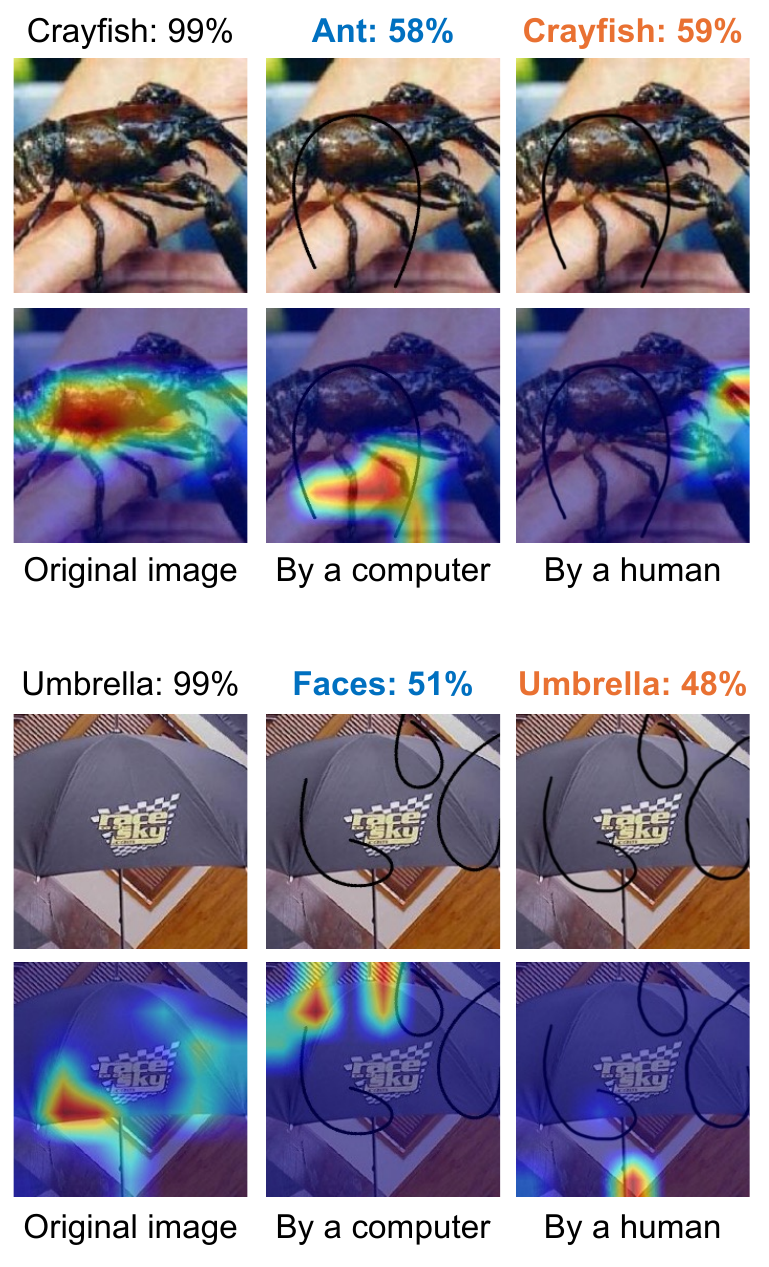}
        \subcaption{Failure cases}
        \label{fig:gradcam:failure}
    \end{minipage}
    \caption{Examples of GradCam analysis. Each visualization result shows where the ResNet-50 classifier focuses on.}
    \label{fig:gradcam}
\end{figure}

We find that the misalignment between computer-generated and human-replicated attacks sometimes largely changes the area of focus of the classifier. \cref{fig:gradcam} shows examples of GradCAM analysis results for the ResNet-50 classifier. \cref{fig:gradcam:success} shows the cases where both computer-generated and human-replicated attacks successfully fool the ResNet-50 classifier. In both cases in \cref{fig:gradcam:success}, the area of focus is approximately the same when we add a computer-generated attack and a human-replicated attack. \cref{fig:gradcam:failure} shows the cases where computer-generated attacks successfully fool the classifier, but human-replicated attacks fail. The area of focus is different when a human replicates the computer-generated attack in both cases in \cref{fig:gradcam:failure}. In many cases where human-replicated attacks fail, the classifier's area of focus changes drastically when a human replicates the computer-generated attack as shown in \cref{fig:gradcam:failure}.

Also, some GradCam analysis results help us understand why adversarial doodles lead to misclassification. In the first example of \cref{fig:gradcam:success}, both computer-generated and human-replicated adversarial doodles make the classifier classify a crab image as an ant. When the classifier classifies the original image, it focuses on the lower-left part of the crab. The computer-generated and human-replicated adversarial doodles hide this part, so the classifier focuses on one leg of the crab. As the leg looks like an ant's leg, the classifier classifies each attacked image as an ant.

\subsection{Transferability}

We evaluate adversarial doodles' transferability: how adversarial doodles made to fool a certain classifier (source classifier) can also fool other classifiers (target classifier). We use the same experimental conditions as \cref{subsec:random-affine}, and calculate how human-replicated adversarial doodles for the ResNet-50 classifier can also fool ViT-B/32. Here, we can view the ResNet-50 as the source classifier, and ViT-B/32 as the target classifier. First, we collect images where the attack successfully fools the source classifier. Second, we exclude images that the target classifier fails to classify even when we add no attack. At this step, we denote the number of filtered images as $N_{\mathrm{total}}$. Finally, we input the attacked images to the target classifier and confirm if the target classifier is fooled. We denote the number of cases where the target classifier misclassifies the attacked image as $N_{\mathrm{success}}$. Then we calculate the attack transferability score as $\frac{N_{\mathrm{success}}}{N_{\mathrm{total}}}$. We also evaluate the reverse experimental settings: We view ViT-B/32 as the source classifier and ResNet-50 as the target classifier and calculate the transferability score in the same way. 

\cref{tab:transferability} shows the results. Adversarial doodles are not transferable: the attack transferability scores are less than $0.3$ in every condition. However, random affine transformation leads to more transferable attacks.

\begin{table}[tb]
    \caption{Attack transferability scores.}
    \label{tab:transferability}
    \centering
    \begin{tabular}{@{}lllll@{}} \toprule
         Source & Target   & $L$ & Random trans. & score \\ \midrule
        ResNet-50 &ViT-B/32& $1$          &                & $0.14$ \\ \cmidrule(){4-5}
            &    &              &     \checkmark & $\bm{0.27}$ \\ \cmidrule(){3-5}
            &     & $3$        &                 & $0.13$ \\ \cmidrule(){4-5}
            &     &          &              & $\bm{0.24}$ \\ \midrule
       ViT-B/32  & ResNet-50 & $1$          &                & $0.0$ \\ \cmidrule(){4-5}
            &    &              &     \checkmark & $\bm{0.17}$ \\ \cmidrule(){3-5}
            &     & $3$        &                 & $0.11$ \\ \cmidrule(){4-5}
            &     &          &      \checkmark & $\bm{0.20}$ \\ \bottomrule
    \end{tabular}
\end{table}

\section{Limitation and Future Work}

Optimizing color could provide further insights into DNN classifiers. We expect that colors are one of the most essential elements when DNNs classify images. Therefore, we will find additional intriguing and describable insights when we optimize \bez curves' color.

We must add additional techniques to generate attacks that are less noticeable to the human eye for the practical attackers' scenario. Our proposed approach sometimes results in attacks that destroy images' semantics. In future work, we should consider this issue when aiming at the practical usage of adversarial doodles.

Moreover, applying adversarial doodles to the physical adversarial attack settings will open a new door to this field. To our knowledge, no experimental research has adopted hand-drawn strokes as a physical adversarial attack method. Adversarial doodles, with their flexible shapes, have the potential to induce misclassification more efficiently and less conspicuously than patches~\cite{roadsigns17} or optical attacks such as laser beams~\cite{advlb} or shadows~\cite{shadows-can-be-dangerous}.

\section{Conclusion}

In this paper, we proposed a novel type of attack called adversarial doodles, which have interpretable and human-drawable shapes. We optimized a set of control points of \bez curves in a gradient-based way, utilizing the differentiable rasterizer with random affine transformation and regularization of the doodled area. We empirically showed that adversarial doodles fooled the ResNet-50 and ViT-B/32 classifiers even when replicated by humans. We found describable insights into the relationship between a human-drawn doodle's shape and the classifier's output. Exploring such features of human-drawn doodles as adversarial attacks would open a new door for both attackers and defenders of adversarial attacks.

{\small
\bibliographystyle{ieee_fullname}
\bibliography{egbib}
}

\clearpage

\newcommand\beginsupplement{%
        \setcounter{table}{0}
        \renewcommand{\thetable}{\Alph{table}}%
        \setcounter{figure}{0}
        \renewcommand{\thefigure}{\Alph{figure}}%
        \setcounter{section}{0}
        \renewcommand{\thesection}{\Alph{section}}
     }
\beginsupplement

\part*{Supplementary Material}

\section{Experimental Details}

In this section, we describe the details of our experiments \cref{sec:experiments}. We implement our proposed algorithm with PyTorch~\citeappx{pytorch}. For the differentiable rasterizer~\citeappx{differentiable-vector-graphic}, we use its PyTorch implementation\footnote{\url{https://github.com/BachiLi/diffvg}}.

\paragraph{Classifiers' Training}
When we prepare the ResNet-50 and ViT-B/32 classifiers, we use the pre-trained models as base models. For ResNet-50, we use pre-trained model of BYOL~\citeappx{boyl} and fine-tune only the final layer with Caltech-101~\citeappx{caltech-101} dataset. For ViT-B/32, we use ImageNet-1K~\citeappx{imagenet} pre-trained model, and fine-tune all its layers with Caltech-101. For the optimizer, we adopt Adam~\cite{adam} and set the learning rate to $0.001$.

\paragraph{Attack Optimization}
When we generate an adversarial doodle as described in \cref{alg:optimization}, we try up to three different initial values $\bm{V}$. As Li \etal~\citeappx{differentiable-vector-graphic} point out, the differentiable rasterizer is prone to local optima. Therefore, if we cannot fool the classifier after updating $\bm{V}$ $N_{\mathrm{itr}}$ times in \cref{alg:optimization}, we return to initialize $\bm{V}$ (L4) again. We set the maximum number of trials to three.

When we set $N_{\mathrm{itr}}=10000$ and $B=10$, it takes about $30$ minutes to run the algorithm in \cref{alg:optimization} on Tesla V100 GPU.

\paragraph{Human Subjects and Doodling Tools}
\cref{tab:human-subject-doodling-tools} shows the ages of human subjects we recruit and the doodling tools they use when replicating computer-generated adversarial doodles in \cref{sec:experiments,subsec:random-affine}. We select human subjects from a wide range of age groups. We instruct the human subjects to use their own devices to replicate the attacks. The most commonly used device is an image editing application called Microsoft paint\footnote{A basic drawing and image-editing application that comes pre-installed on Windows.}.

\begin{table}[tb]
    \centering
    \caption{The ages of human subjects and doodling tools they use.}
    \label{tab:human-subject-doodling-tools}
    \begin{tabular}{@{}lll@{}} \toprule
        ID & Tool & Age \\ \midrule
         $1$ & Preview\tablefootnote{An application pre-install on macOS} & $36$ \\
         $2$ & Microsoft paint & $48$ \\
         $3$ & ibisPaint X\tablefootnote{The new version of ibisPaint.} & $39$ \\
         $4$ & Microsoft paint & $54$ \\
         $5$ & Microsoft paint & $32$ \\
         $6$ & Preview & $38$ \\
         $7$ & ibisPaint\tablefootnote{\url{https://ibispaint.com/}} & $54$ \\
         $8$ & Microsoft paint & $37$ \\
         $9$ & Microsoft paint & $40$ \\
         $10$ & Microsoft paint & $43$ \\
         $11$ & Microsoft paint & $57$ \\
         $12$ & iPad photo app & $47$ \\
         $13$ & Paint 3D\tablefootnote{\url{https://apps.microsoft.com/detail/9nblggh5fv99}} & $44$ \\
         $14$ & FireAlpaca64\tablefootnote{\url{https://firealpaca.com/}} & $37$ \\
         $15$ & iPad memo app\tablefootnote{A built-in application of iPads.} & $45$ \\
         $16$ & Microsoft paint & $47$ \\
         $17$ & Microsoft paint & $49$ \\
         $18$ & Microsoft paint & $47$ \\
         $19$ & iPad markup\footnote{A built-in feature of iPad.} & $47$ \\
         $20$ & Microsoft paint & $68$ \\ \bottomrule
    \end{tabular}
\end{table}

\section{Examples of Attacks for ViT-B/32}

\begin{figure}[tb]
    \centering
    \begin{minipage}{0.98\hsize}
        \begin{minipage}{\hsize}
            \centering
            \includegraphics[width=\hsize]{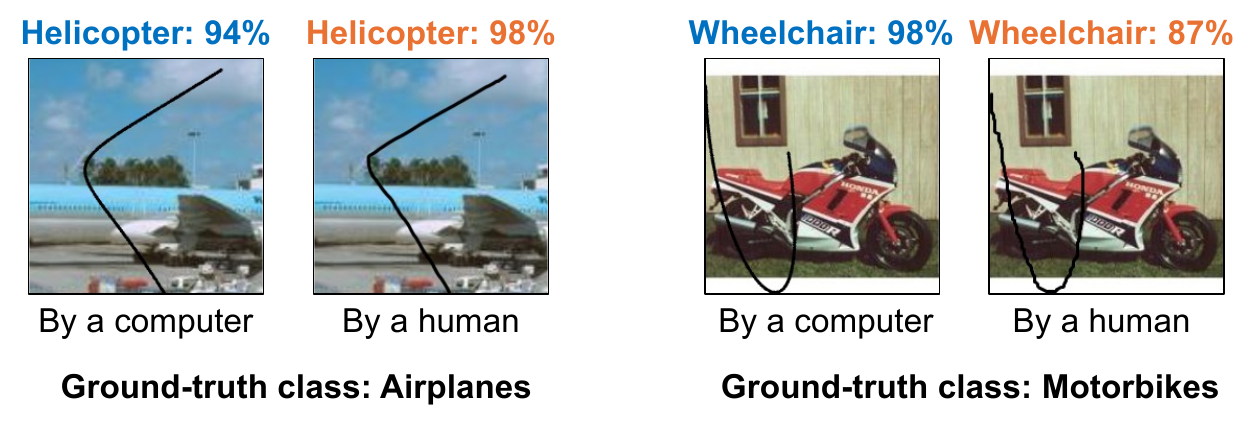}
        \end{minipage}
        \begin{minipage}{\hsize}
            \centering
            \includegraphics[width=\hsize]{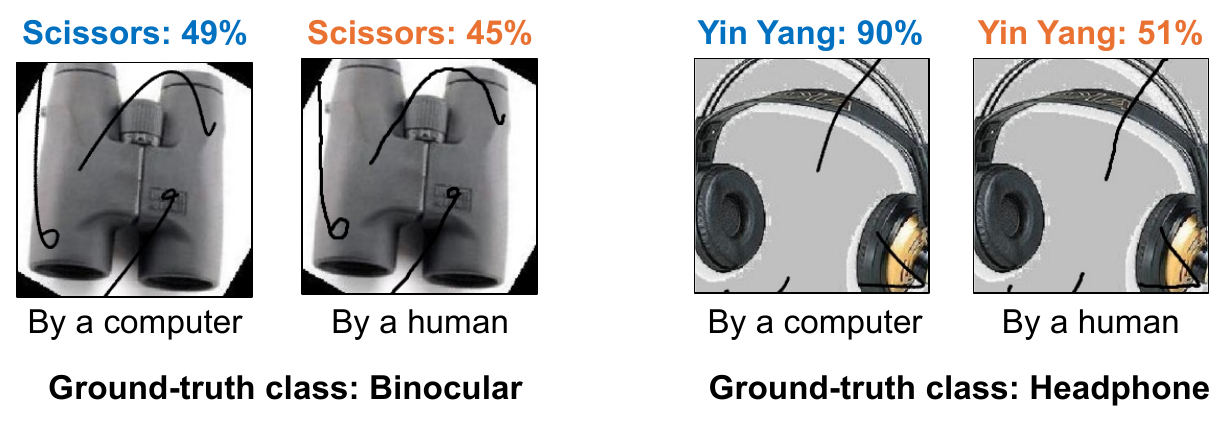}
        \end{minipage}
        \subcaption{Success cases}
        \label{fig:replicated-attack-examples:success}
    \end{minipage}
    \begin{minipage}{0.98\hsize}
        \begin{minipage}{\hsize}
            \centering
            \includegraphics[width=\hsize]{figures/supplementary/3_bez3_success_vit.pdf}
        \end{minipage}
        \begin{minipage}{\hsize}
            \centering
            \includegraphics[width=\hsize]{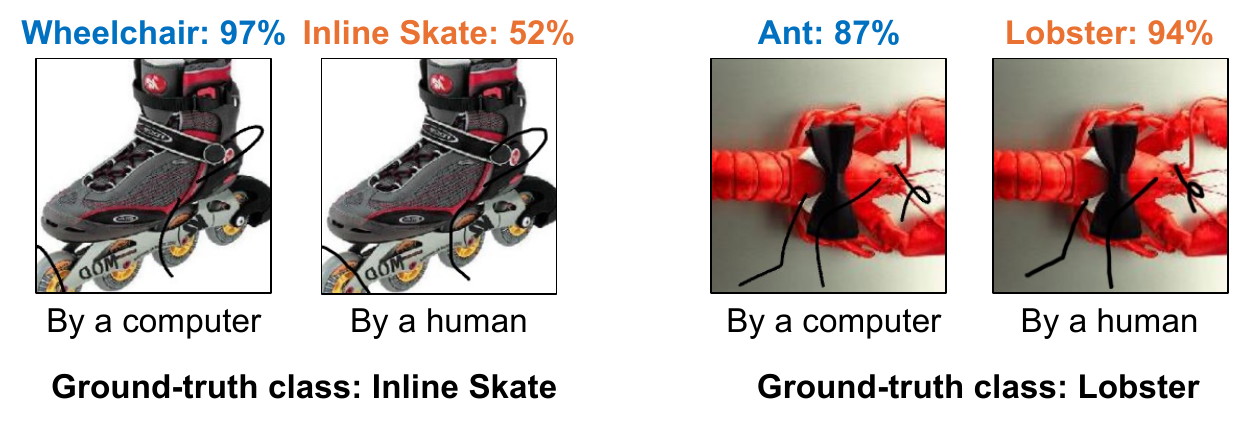}
        \end{minipage}
        \subcaption{Failure cases}
        \label{fig:replicated-attack-examples:failure}
    \end{minipage}
    \caption{Examples of adversarial doodles for the ViT-B/32 classifier. \cref{fig:replicated-attack-examples:success} shows the cases where both computer-generated and human-replicated attacks successfully fool the ViT-B/32 classifier. \cref{fig:replicated-attack-examples:failure} shows the cases where computer-generated attacks successfully fool the classifier, but human-replicated attacks fail.}
    \label{fig:replicated-attack-examples-vit}
\end{figure}

\begin{figure}[tb]
    \centering
    \begin{minipage}[tb]{0.48\hsize}
        \centering
        \includegraphics[width=\hsize]{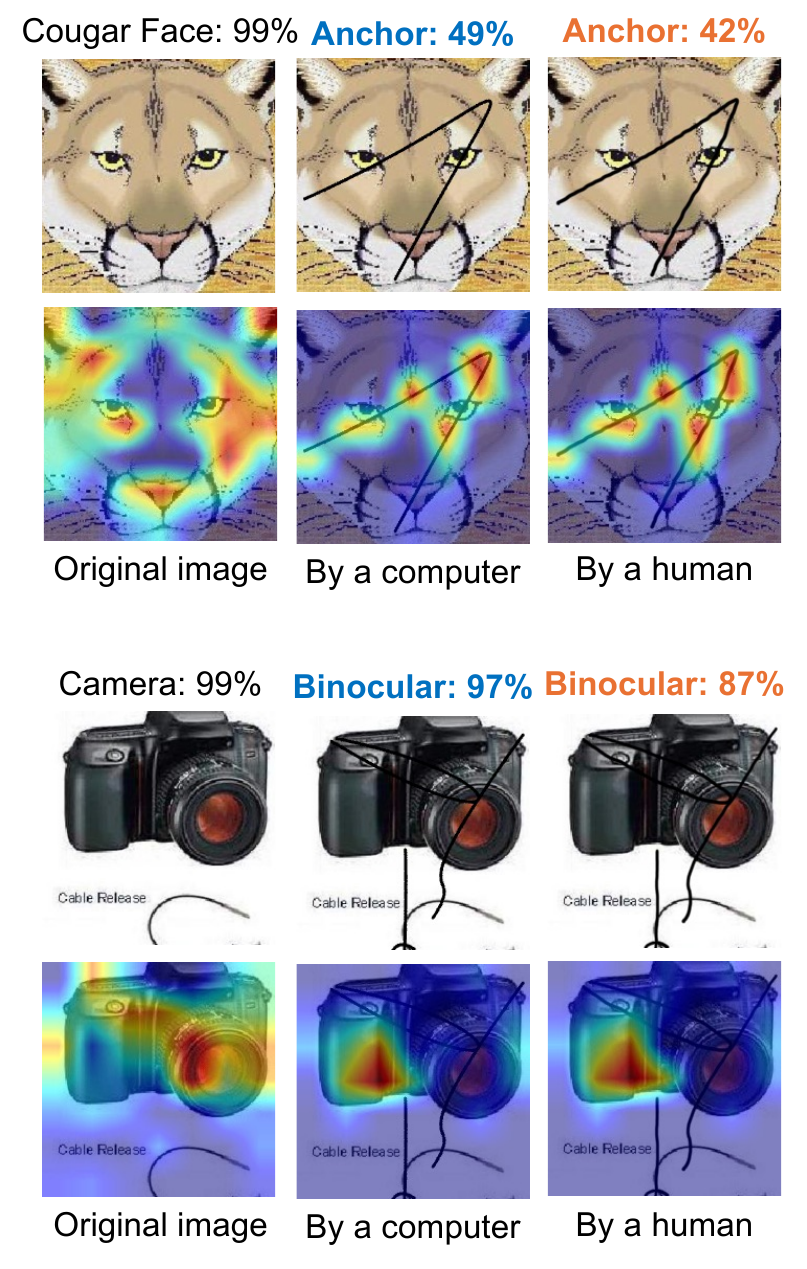}
        \subcaption{Success cases}
        \label{fig:vit-gradcam:success}
    \end{minipage}
    \hfill
    \begin{minipage}[tb]{0.48\hsize}
        \centering
        \includegraphics[width=\hsize]{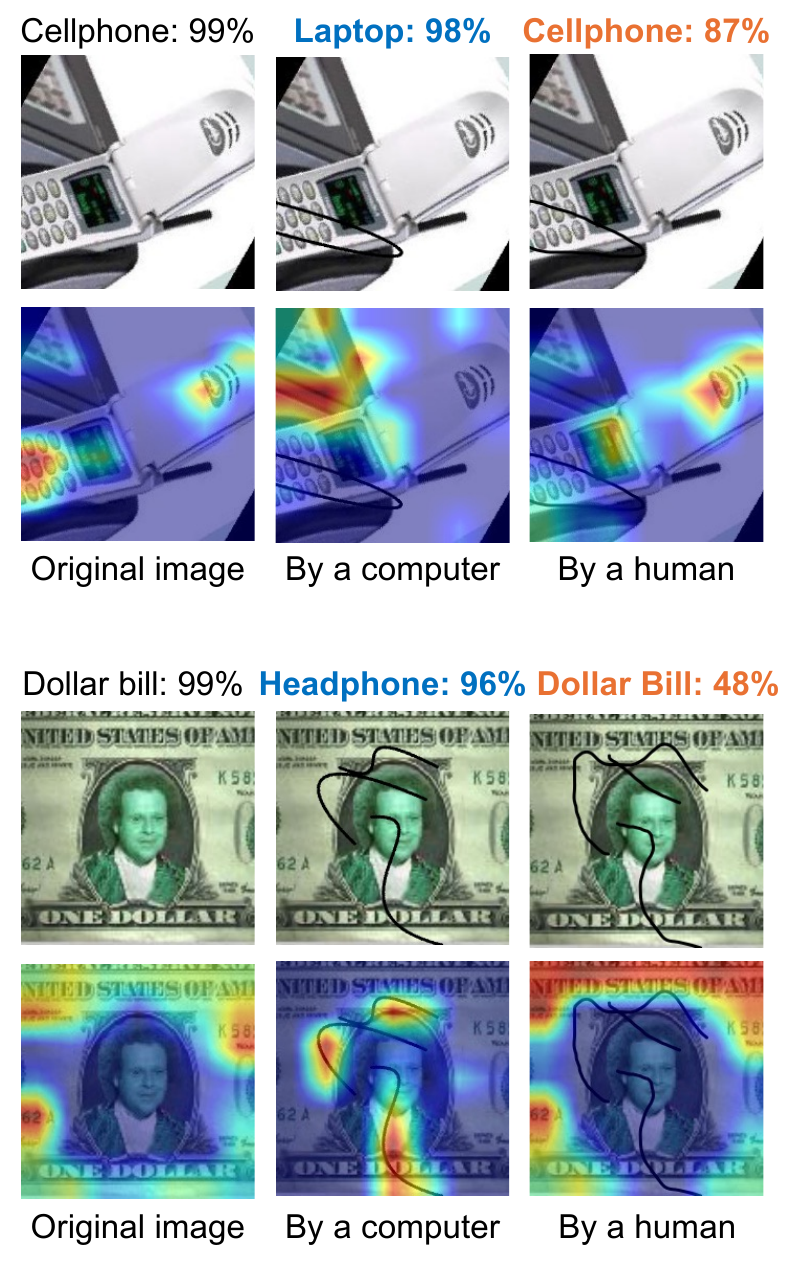}
        \subcaption{Failure cases}
        \label{fig:vit-gradcam:failure}
    \end{minipage}
    \caption{Examples of GradCam analysis. Each visualization result shows where the ViT-B/32 classifier focuses on. \cref{fig:vit-gradcam:success} shows the cases where both computer-generated and human-replicated attacks successfully fool the ViT-B/32 classifier. \cref{fig:vit-gradcam:failure} shows the cases where computer-generated attacks successfully fool the classifier, but human-replicated attacks fail. }
    \label{fig:vit-gradcam}
\end{figure}

We provide the examples of adversarial doodles for the ViT-B/32 classifier in \cref{fig:replicated-attack-examples-vit} and the GradCam analysis results in \cref{fig:vit-gradcam}. We omit these samples in the main paper because of space limitation.

\small{
\bibliographystyleappx{ieee_fullname}
\bibliographyappx{sup}
}

\end{document}